\def\year{2022}\relax
\documentclass[letterpaper]{article} 
\usepackage{aaai22}  
\usepackage{times}  
\usepackage{helvet}  
\usepackage{courier}  
\usepackage[hyphens]{url}  
\usepackage{graphicx} 
\usepackage{natbib}  
\usepackage{caption} 
\DeclareCaptionStyle{ruled}{labelfont=normalfont,labelsep=colon,strut=off} 
\usepackage{algorithm}
\usepackage{algorithmic}

\usepackage{CJKutf8}
\usepackage[utf8]{inputenc}
\usepackage{natbib}
\usepackage{graphicx}
\usepackage{xcolor}
\usepackage[algo2e]{algorithm2e}
\usepackage{array}

\usepackage[namelimits]{amsmath} 
\usepackage{amssymb}             
\usepackage{amsfonts}            
\usepackage{mathrsfs}            
\usepackage{amsopn}
\usepackage{booktabs}
\usepackage{multirow}
\usepackage{multicol}
\graphicspath{{figures/}}

\usepackage{amsopn}
\usepackage{booktabs}
\usepackage{multirow}
\usepackage{etoolbox}
\usepackage{textpos} 

\usepackage{newfloat}
\usepackage{listings}
\floatstyle{ruled}
\newfloat{listing}{tb}{lst}{}
\floatname{listing}{Listing}
\title{EfficientCLIP: Efficient Cross-Modal Pre-training by Ensemble Confident Learning  and Language Modeling}
\author{
 Jue Wang$^{\dag\ddag}$, Haofan Wang$^{\ddag}$, Jincan Deng$^{\ddag}$,  Weijia Wu$^{\ddag\P}$, Debing Zhang$^{\ddag}$\\
 $^\dag$Zhongnan University of Economics and Law, $^\P$Zhejiang University, $^\ddag$Kuaishou Technology \\
 {\tt\small \ 201821090281@stu.zuel.edu.cn, \{wanghaofan, dengjincan\}@kuaishohu.com, weijiawu@zju.edu.cn\, , zhangdebing@kuaishohu.com}
}

\usepackage{bibentry}

\begin{document}

\maketitle

\begin{abstract}
While large scale pre-training has achieved great achievements in bridging the gap between vision and language, it still faces several challenges. First, the cost for pre-training is expensive. Second, there is no efficient way to handle the data noise which degrades model performance. Third, previous methods only leverage limited image-text paired data, while ignoring richer single-modal data, which may result in poor generalization to single-modal downstream tasks. In this work, we propose an EfficientCLIP method via Ensemble Confident Learning to obtain a less noisy data subset. Extra rich non-paired single-modal text data is used for boosting the generalization of text branch. We achieve the state-of-the-art performance on Chinese cross-modal retrieval tasks with only 1/10 training resources compared to 
CLIP and WenLan, while showing excellent generalization to single-modal tasks including text retrieval and text classification.
\end{abstract}

\section{Introduction}
Pre-training has achieved great progress in Natural Language Processing~\cite{brown2020language,devlin2018BERT,lewis2019bart,liu2019roBERTa,radford2019language} and Computer Vision~\cite{chen2020simple,dosovitskiy2020image,kolesnikov2019big} tasks. Recently, multi-modal pre-training~\cite{huo2021wenlan,radford2021learning,jia2021scaling} attracts widespread attention, where large scale weakly correlated multi-modal data on the internet is utilized to learn cross-modal representation by contrastive learning~\cite{hadsell2006dimensionality}. However, existing multi-modal pre-training methods face several challenges. First, as the scale of data expands, pre-training requires expensive training resources, CLIP~\cite{radford2021learning} costs 3584 GPU-days and WenLan~\cite{huo2021wenlan} costs 896 GPU-days both on NVIDIA A100. Second, the raw internet data is noisy (as shown in Appendix, Figure \ref{more case}), which wastes training resources and extremely degrades the performance of model~\cite{algan2021image,carlini2021poisoning,northcutt2021confident,shen2020noise}. Third, previous multi-modal pre-training methods only use limited image-text pairs, while ignoring richer single-modal text data, which results in poor generalization to many downstream NLP tasks and scenes~\cite{li2020unimo}.

\begin{figure}[t]
\centering
\includegraphics[width=0.5\textwidth]{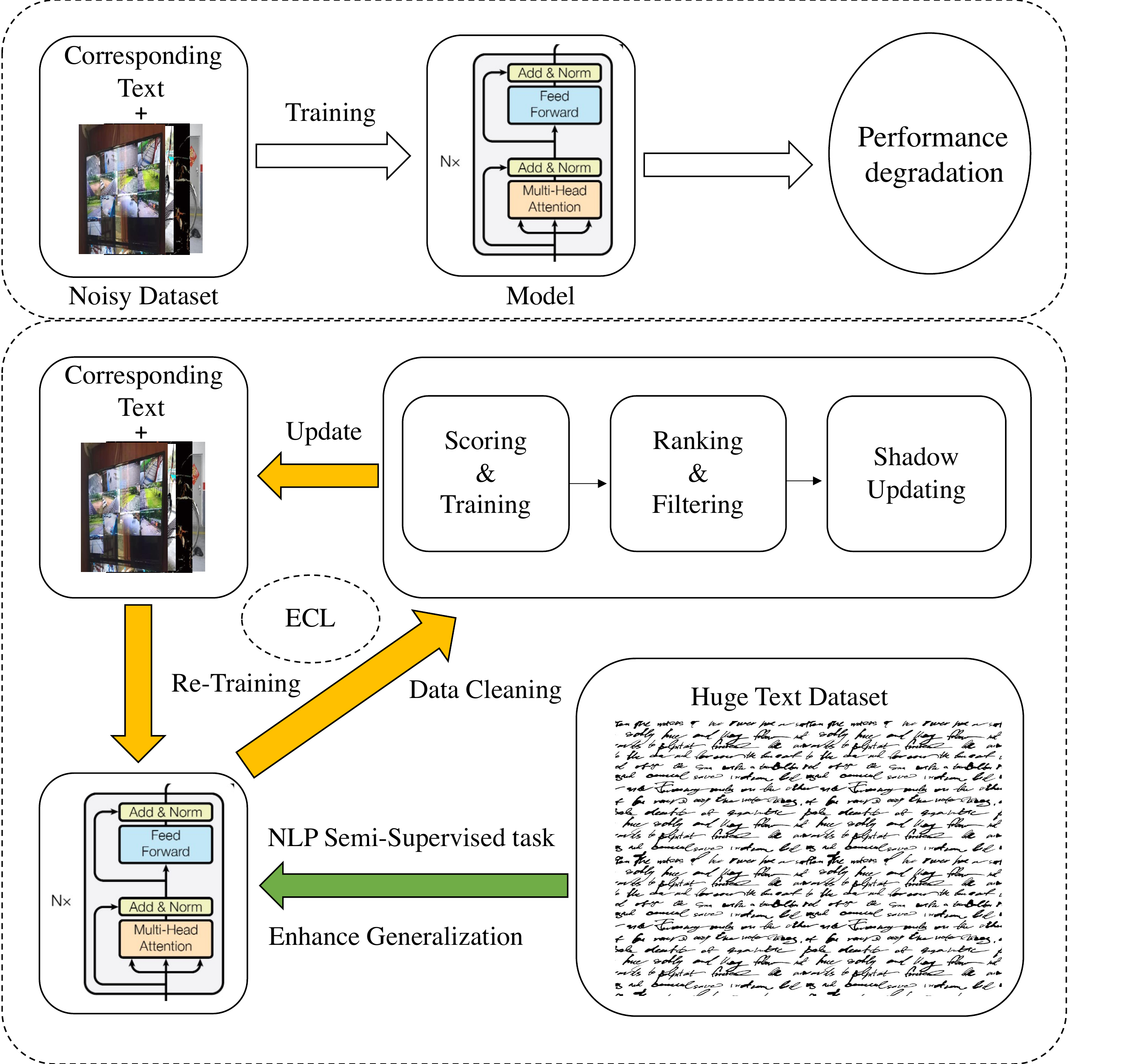}
\vspace{-6mm}
\caption{\textbf{Comparison of training processes.} The flowchart above shows a common procedure of previous cross-modal pre-training methods. The flowchart below shows our difference, where the training dataset is progressively updated via an Ensemble Confident Learning (ECL) strategy.}
\label{fig:comparison}
\vspace{-6mm}
\end{figure}

There are numerous works~\cite{northcutt2021confident,shen2020noise,xie2020self} that contribute to training noise-robust models or developing noise-free strategies for vision tasks. In contrast, few studies have focused on handling noise in cross-modal pre-training. Inspired by confident learning~\cite{northcutt2021confident}, we propose a novel method EfficientCLIP via an Ensemble Confident Learning (ECL) strategy, where models at different training epochs are ensembled to estimate the data distribution through several continuous iterative steps. We show that ECL strategy helps model  save training resources and speeds up convergence by building a smaller subset with less noisy data for training. To further boost the generalization performance on downstream tasks and scenes, motivated by the success of self-supervised tasks~\cite{devlin2018BERT,dong2019unified,lewis2019bart,gao2021simcse} in NLP, we use extra non-paired single-modal data which is available in much richer scenes than cross-modal data, specifically text data in our case, to enhance the text encoder via self-supervised learning tasks. For the purpose of saving the cost of training to the greatest extent, our image encoder is built on the top of CLIP~\cite{radford2021learning} because of its robustness validated by ~\cite{shen2021much}. To make up for the scarcity of the open Chinese multi-modal model and solve various Chinese application scenarios, our model is trained with data in Chinese. The key difference of the training process between our method and previous works can be found in Figure \ref{fig:comparison}.

We evaluate the performance on both cross-modal and single-modal tasks and achieve state-of-the-art (SOTA) results on corresponding benchmarks. For instance, with only 1/10 training resources compared to WenLan (SOTA in Chinese) and CLIP, EfficientCLIP outperforms WenLan~\cite{huo2021wenlan} on Recall@1 (R@1) and Recall@5 (R@5) by 3.6\% and 5.9\% respectively, on the cross-modal retrieval tasks of AIC-ICC~\cite{wu2019large} (the largest Chinese caption dataset). We also exceed CLIP~\cite{radford2021learning} on R@1 and R@5 by 4.39\% and 5.11\% on COCO~\cite{veit2016coco} text-to-image retrieval task. 
 Moreover, enriched by extra single-modal data, EfficientCLIP also shows great generalization on single-modal tasks. We outperform the SOTA on text retrieval task and exceed benchmarks with a large margin on text classification task. All experiments are conducted on zero-shot setting, in addition to the text classification task. Details can be found in Sec \ref{comparison}. Our key contributions are summarized as below:

1. We propose EfficientCLIP via an \textbf{Ensemble Confident Learning (ECL)} strategy for efficient cross-modal pre-training, which pruning noisy data adaptively.

2. We achieve the state-of-the-art on cross-modal retrieval tasks with only \textbf{1/10 training resources} compared to benchmarks such as CLIP and WenLan.

3. We show the value of \textbf{non-paired single-modal data} on cross-modal pre-training, and achieve great generalization ability to downstream single-modal tasks.

\section{Related Work}
\subsection{Contrastive Learning}
Contrastive learning~\cite{hadsell2006dimensionality,le2020contrastive} is a kind of representation learning by contrasting positive pairs against negative pairs. Recent literatures~\cite{chen2020simple,gao2021simcse,he2020momentum} have achieved great success in both representation learning and unsupervised learning tasks by contrastive learning. Among them, SimCLR~\cite{chen2020simple} proposes to match positive sample pairs by various data augmentations, and uses larger batch size with richer negative samples for contrastive learning to get a robust visual representation. MOCO~\cite{he2020momentum} uses the dictionary as a queue to solve the dependence of contrastive learning on large batch size. In the field of NLP, SimCSE~\cite{gao2021simcse} proposes to use dropout to make the same text for mini data augmentation as a positive sample pair, and uses different texts as negative samples. 
%
%
%

\subsection{Two-Tower Structure Models}

Two-tower structure models trained on large scale web datasets have been successfully used in multi-modal tasks and achieve excellent results on many downstream tasks.  
In two-tower structures, visual and language information are encoded independently without any fusion via contrastive learning scheme for efficient discrimination. 
CLIP~\cite{radford2021learning} trains with 400 million image-text data pairs from the internet after simple data cleaning, it achieves excellent performance on image-text retrieval tasks and zero-shot image recognition tasks. ALIGN~\cite{jia2021scaling} further expands the scale of image-text paired data, and uses 1 billion data to train without any cleaning, indicating that the expansion of the data scale can suppress the influence of noise for the model to 
some extent. WenLan~\cite{huo2021wenlan} trains with Chinese paired data and achieves the best performance on image-text retrieval task in the Chinese scene.

\begin{figure*}[t]
\begin{center}
\includegraphics[width=1\textwidth]{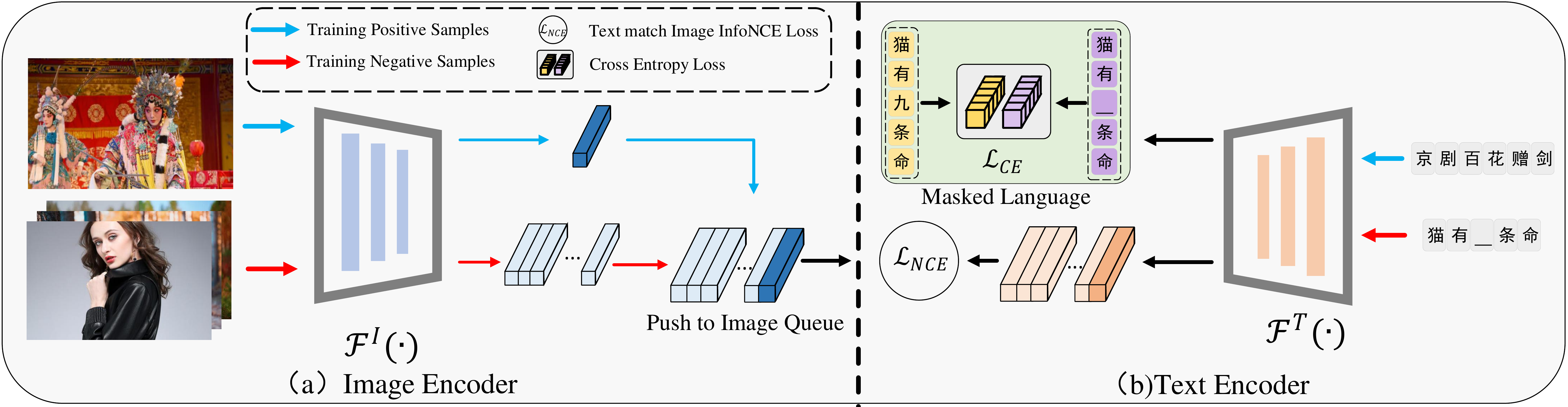}
\vspace{-6mm}
\caption{\textbf{Illustration of the whole pipeline.} (a) Image Encoder: Image features are pushed to the image queue (see Appendix \ref{memory} for details of image queue). (b) Text Encoder: Text features are obtained by text encoder for contrastive learning with image queue. Notation: $F^I(.)$ and $F^T(.)$ denote image encoder and text encoder~\cite{dosovitskiy2020image}, respectively.}

\label{fig:cost}
\end{center}
\vspace{-8mm}
\end{figure*}

\subsection{Learning with Noisy Data}

There is a continued interest in community on training models directly from enormous and low-cost web data. 
However, a large scale of noisy data existing on the internet  cause negative effects for model training. 
To utilize numerous and low-cost web data, many researchers attempt to explore noise-robust methods~\cite{chen2015webly,joulin2016learning,fergus2005learning,schroff2010harvesting} for training models.
~\citet{fergus2005learning} exploit images from the Google search engine for image categorization based on an improved pLSA method.
~\citet{schroff2010harvesting} propose an approach to automatically harvest images on the internet for learning visual recognition classifiers.
~\citet{krause2016unreasonable} show that models learning from a large scale of web data outperform those learning from a limited amount of human-annotated dataset for classification tasks.
Confident Learning~\cite{northcutt2021confident} achieves effective noise filtering through three continuous iterative steps of Count, Rank and Prune, and Re-Training, and trains a more robust model.
%
%

However, these works are only designed for common object detection and classification tasks, which cannot directly be applied effectively to cross-modal pre-training tasks.
As far as we know, current multi-modal pre-training methods still lack an effective training method to effectively learn from noisy data.
Thus, in this paper, we propose an Ensemble Confident Learning (ECL) strategy to fill this gap.

\section{Methodology}

In this section, we illustrate the steps of our efficient training strategy. In Sec 3.1, we first introduce how we transfer CLIP's knowledge from English domain to Chinese domain. From Sec 3.2 to Sec 3.3, we demonstrate two core operations, Ensemble Confident Learning and extra single-modal self-supervised learning respectively. We also describe a memory queue method based on the idea of dictionary as a queue~\cite{he2020momentum} and the pseudocode of our approach, which can be found in Appendix \ref{memory} and Figure \ref{pseudocode} in Appendix, respectively. The pipeline is shown in Figure \ref{fig:cost}.

\subsection{Cross-language Knowledge Distillation}

It is time-costing and expensive to train a cross-modal pre-trained model from scratch. To conveniently obtain a text encoder in Chinese domain, we perform cross-language knowledge distillation.
A transformer-based Chinese text encoder (refer to Appendix, Table \ref{hyperparameters} for detailed structure) is initialized as a student model and distills knowledge from a teacher model (CLIP's text encoder is adopted).

\begin{equation}
    Loss= \sum_{i=1}^{N} (F^C(T_c^{(i)})-F^{E}(T_e^{(i)}))^2.
\end{equation}

\begin{figure}[h!]
\centering
\includegraphics[width=0.5\textwidth]{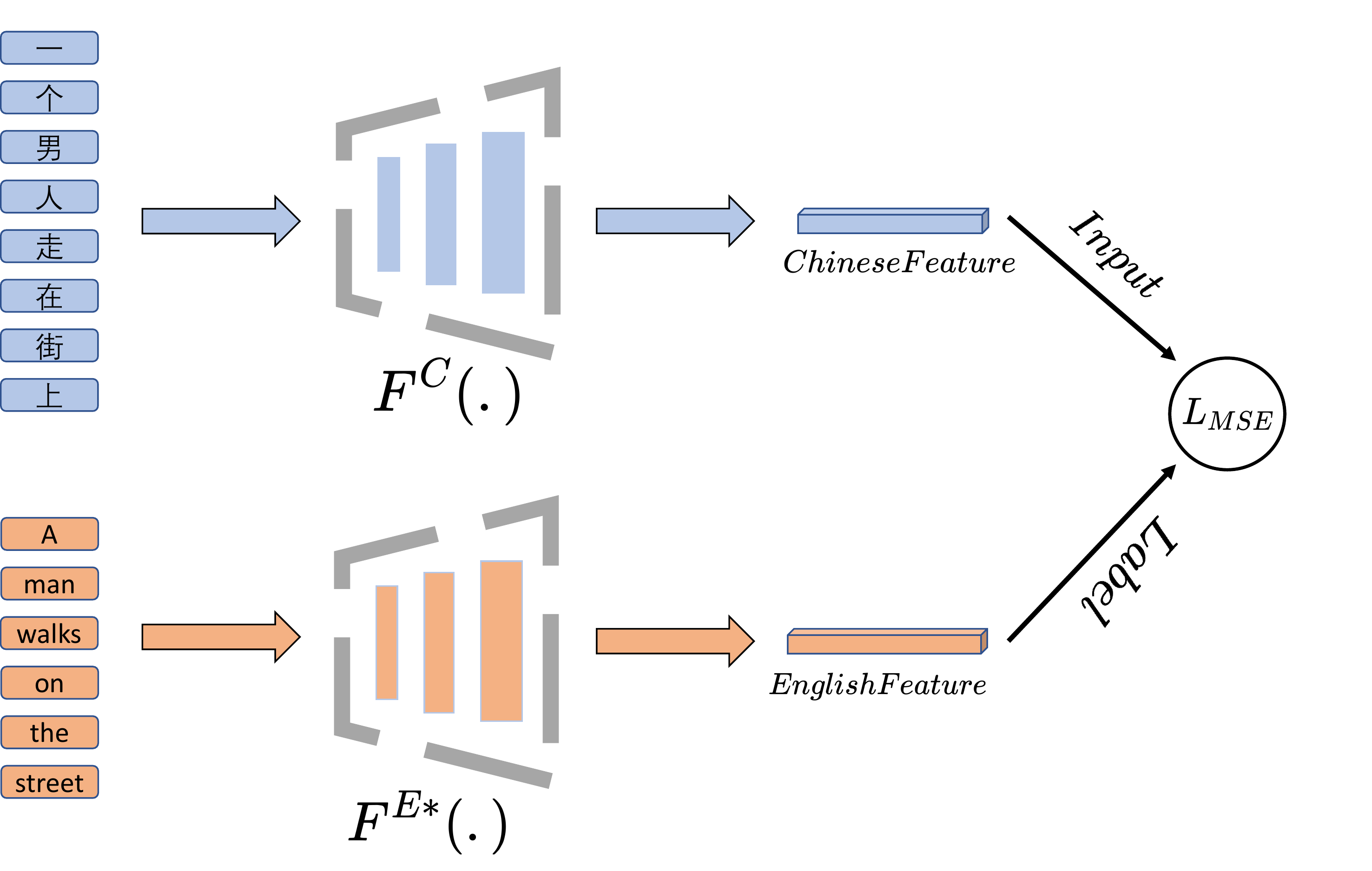}
\vspace{-6mm}
\caption{\textbf{Cross-language distillation.} $F^C$ represents the Chinese Encoder, $F^{E}$ represents the English Encoder. The frozen model is indicated by *. Our goal is to minimize the distance between their outputs via an MSE loss.}
\label{fig:Distillation Model}
\vspace{-6mm}
\end{figure}

The Chinese-English pairs are defined as $T_{ce}=\{(T_c^{(i)}, T_e^{(i)}), i \in [1, N]\}$, where $T_c^ {(i)}$ and $T_e^{(i)}$ represent Chinese and English texts, respectively. Figure  \ref{fig:Distillation Model} shows the structure of distillation model, we freeze the parameters of the English teacher model $F^E$, and only update the Chinese student model $F^C$ to minimize the distance between their outputs via an MSE loss formulated as above.

\subsection{Ensemble Confident Learning}

Large scale image-text datasets crawled from the internet have been widely used in pre-training. As indicated by ~\cite{carlini2021poisoning,northcutt2021confident,shen2020noise}, excessive noisy data negatively impacts the model's performance and training efficiency. ALIGN~\cite{jia2021scaling} and WenLan~\cite{huo2021wenlan} demonstrate that the large-scale pre-training with expensive resources can suppress the influence of noise to some extent, but these training resources are usually not available for general researchers. An alternative is to establish cleaned datasets like COCO~\cite{veit2016coco}, or Conceptual Captions ~\cite{sharma2018conceptual}. However, as the high-quality manual annotations or complex treatments are needed, they are often limited by their scales, which result in the poor generalization. Driven by these obstacles, a compromised way based on the idea of Confident Learning ~\cite{northcutt2021confident} named as Ensemble Confident Learning (ECL) is designed. Instead of removing all noisy pairs at once, ECL strategy adopts the same way as Confident Learning and adaptively remove noisy data from the training set, as the distribution of dataset is hard to estimate.


As the model generally performs more discriminative on those high-correlated pairs (related experiments can be found in Appendix \ref{exp of ecl}), we propose to adaptively and iteratively remove the noisy pairs (low-correlated) by means of the discriminative ability of model to data distribution. First, we use the distillation model as initialization which is already equipped with the basic discriminative powers (it has been proved in Appendix \ref{exp of ecl}), and additionally establish a scoring shadow model that only updates parameters at the beginning of each epoch. 
For the $K^{th}$ epoch, we define the dataset as $D_K=\{d_1,d_2...d_i...d_n\}$ where $d$ is an image-text pair, the score gained through the scoring shadow model ($S_K$). The pre-trained model which keeps updating is denoted as $T$. ECL strategy consists of three steps: 

\textbf{(1) Scoring \& Training:}. For each image-text pair $d_i$, we use the scoring shadow model to calculate its correlation score $S_K(d_i)$ at the current epoch, and update the total score $C_{K+1}(d_i)$ based on the exponential smoothing algorithm: 
\begin{equation}
C_{K+1}(d_i)=\alpha*C_K(d_i) + S_K(d_i),
\end{equation}
where $\alpha$ refers to the smoothing factor. 
In this way, we get the correlation score of each image-text pair, and train the pre-trained model $T$ with contrastive loss at the same time. 

\textbf{(2) Ranking \& Filtering:} Based on the total score $C_K(d_i)$ of each pair, we reorder the dataset $D_K$ in descending order: 
$D_K^*=\{d_1^*,d_2^*, \cdots, d_n^* \}$. 
To filter out the noisy pairs, we set a filtered rank $\lambda$ and retain the training pairs whose correlation ranks before the $\lambda$. The obtained pairs are used as the training dataset at the $(K+1)^{th}$ epoch: 
\begin{equation}
    D_{K+1}=\{d_1^*,d_2^*, \cdots, d_{\lambda \times n}^*\}.
\end{equation}

\textbf{(3) Shadow Updating:} At the beginning of the next epoch, we update the parameters of the scoring shadow model $S_{K+1}$ with the parameters of the pre-trained model $T$. Then, we return to step (1) for the training of next epoch. We analyze the effectiveness of ECL in Appendix \ref{eff of ecl} to show the reason of using the distillation model, and the effect of "ensemble".

\subsection{Masked Language Model}

Existing cross-modal pre-trained models utilize large scale image-text pairs from the internet, while not realizing that those paired datasets are usually scene-limited. We observe that image-text pairs are common in some specific domains, such as Wikipedia, News and several human-annotated public datasets. However, in other more professional domains such as Technology and Medical, paired data is scarce, single-modal non-paired data is abundant instead. Driven by the idea of multimodal few-shot learning~\cite{tsimpoukelli2021multimodal}, we additionally leverage extra single-modal text data from various scenes for self-supervised learning to enhance the generalization of model to downstream tasks.

We adopt the Masked Language Model (MLM) task proposed by BERT~\cite{devlin2018BERT} for self-supervised training. 
MLM takes advantage of bidirectional semantic information, and only requires a simple masking operation on the original text. Given a sequence of text as $\{t_1, t_2, t_3, \cdots, t_n\}$, we randomly replace the words with [mask] token. 
$T$ and $T_{mask}$ represent the unsubstituted and the substituted tokens respectively. The optimization objective of the text encoder is formulated as:

\begin{equation}
    W^*=\underset{W}{\arg\max} P(T_{mask}|T,W),
\end{equation}
where $W$ refers to the parameters of the text encoder. 

We find that cross-modal pre-training can benefit from additional prevalent single-modal non-paired data. 
The reasons can be explained from two aspects, as described next.

First, the model gains more knowledge about rich scenes from additional text data, which helps avoid the problem of limited distribution of image-text pairs.
Second, the MLM task facilitates the model to pay more attention to the relationship between words and helps the model avoid catastrophic forgetting of token-level knowledge, which results in the improvement of transferring tasks~\cite{gao2021simcse}. Ablation study can be found in Sec \ref{nlp pre-train}.

\section{Experiment}
\label{4 Experiment}

We describe the training datasets in Sec 4.1 and implementation details from Sec 4.2 to 4.4 before showing promising results on the public evaluation datasets. We compare with the state-of-the-art in Sec 4.5 and conduct comprehensive ablation studies in Sec 4.6 to exhibit the effect of each module. More results can be found in Appendix.

\subsection{Datasets}

We establish 3 training sets and 1 validation set based on public datasets (including Chinese-English text pairs, Chinese text data) and web-crawled datasets (including Image-Text training set and validation set). The details of datasets can be found in Appendix \ref{Datasets}.

\subsection{Knowledge Distillation}
\label{distillation}

To conduct cross-language knowledge distillation, we build our distillation model on the top of frozen  CLIP ViT-B/32~\cite{radford2021learning}, and add additional transformer layers on the text encoder for learning. Two models with different hyperparameters are constructed as listed in Appendix, Table \ref{hyperparameters}. We train the distilled Chinese text encoder on a Chinese-English text paired dataset of 80 million size with a batch size of 256. Due to the high correlation between Chinese-English data, the convergence speed of knowledge distillation is 2.5-3.5 times faster than regular contrastive learning. In the experiment, we use 24 V100 GPUs training for two days to converge to the local optimal value.

\subsection{Cross-modal Pre-Training}
\label{Pre-Training}

\subsubsection{Pre-training with ECL}\

\begin{figure}[h!]
\centering
\includegraphics[width=0.5\textwidth]{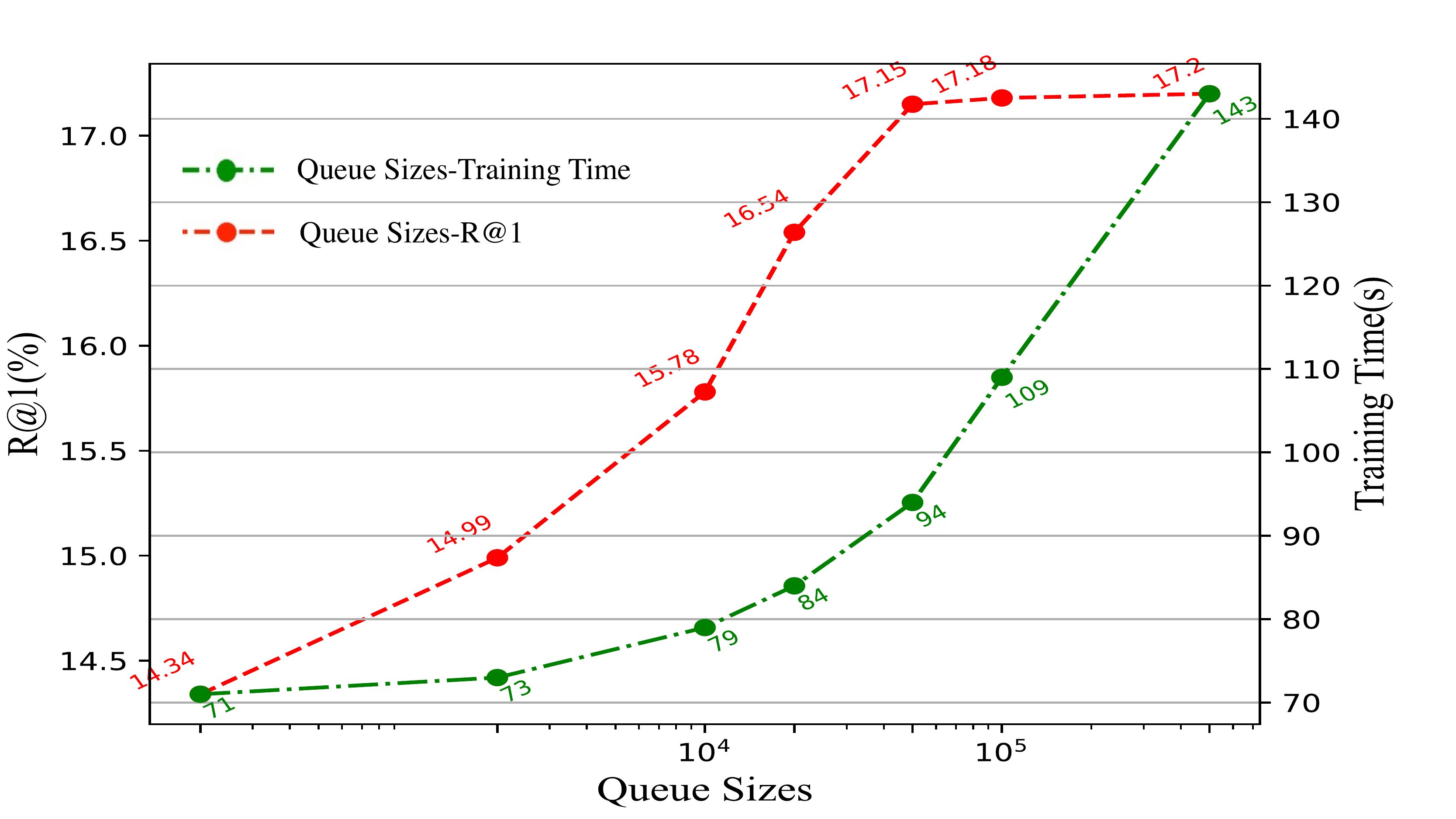}
\vspace{-8mm}
\caption{\textbf{Impact of queue sizes.} The vertical axis is the size of memory queue. R@1 represents the recall 1 score on the text-to-image task of AIC-ICC. Training time represents the training time (second) of models with different queue sizes for training 100 steps.}
\label{queue}
\vspace{-4mm}
\end{figure}

To reduce the cost of pre-training, we use the distillation model as initialization and introduce half-precision optimization based on Deepspeed~\cite{rasley2020deepspeed} framework for efficient distributed training. 
More tricks for speeding up are illustrated in Appendix \ref{speed}.

For the choice of hyperparameters, the similar setting as MOCO~\cite{he2020momentum} and CLIP~\cite{radford2021learning} is adopted: we set the learning rate to 2e-3, weight decay to 1e-4, dropout to the default 0.1, and use cosine warmup schedule as learning rate adjuster. 
To select the optimal size of the queue, we experiment with several parameters including 200, 2,000, 10,000, 20,000, 50,000, 100,000 and 500,000. 
As shown in Figure \ref{queue}, the larger the size of the queue, the higher the accuracy of our model. However, when the query size reaches 50,000, the model accuracy gets saturated and is no longer significantly improved. 
Meanwhile, a too large queue will bring about a great reduction in training efficiency. Considering the trade-off between performance and training efficiency, the queue size is finally set to 50000.

\begin{figure}[h!]
\centering
\includegraphics[width=0.5\textwidth]{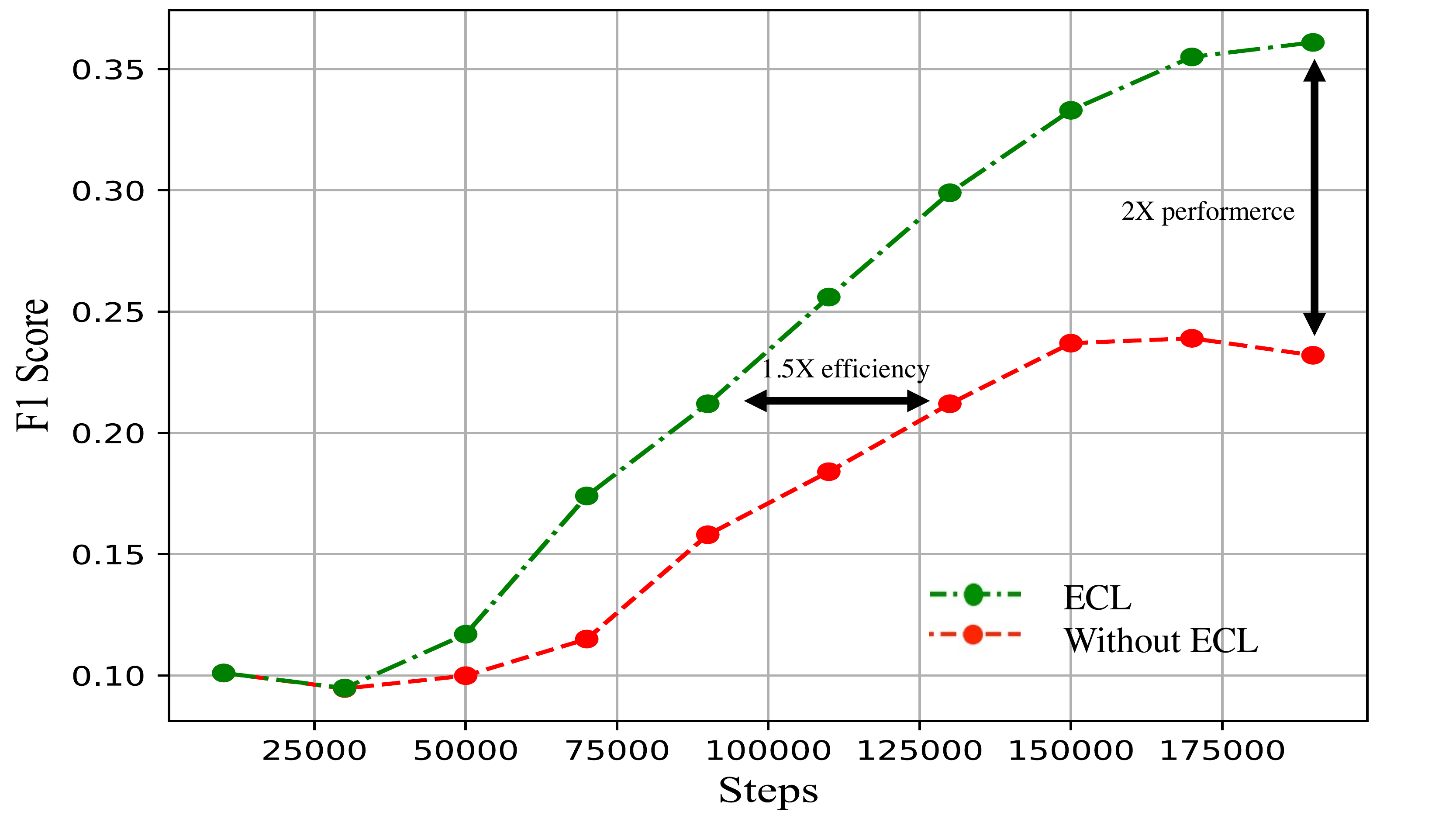}
\vspace{-8mm}
\caption{\textbf{Impact of ECL on performance.} The vertical axis is the training steps, and the horizontal axis is the performance of the f1 score of the training model. As shown, ECL method helps to improve the convergence speed and model performance. }
\label{fig:convergence}
\vspace{-4mm}
\end{figure}

We use the ECL method to filter data in each epoch and evaluate the model's performance on the validation set every 1000 steps. After 9 training epochs, there are only 120 million data left. We find that the score of the validation set does not increase significantly later, thus we stop the data filtering and continue training on the 120 million data.
Benefiting from the effectiveness of adaptively filtering data, we only spend 7 days training our model. 
Compared with the model pre-trained on 300 million data without ECL, our model surpasses it by 8-14\% on the R@1 on the test set of AIC-ICC. The result can be found in Figure \ref{fig:convergence}.

To select a suitable filtered rank for ECL, we experiment with 4 values (0.7, 0.8, 0.9, 0.99), and use  the R@1 of text-to-image task on AIC-ICC for evaluation. 
As illustrated in Table \ref{filtered_values}, when the filtered rank is 0.9, the model reaches the highest R@1 of text-to-image task on AIC-ICC.

We claim that training directly on a large amount of noisy dataset, the model's performance is constrained as shown in Figure \ref{fig:convergence}. With the existence of ECL strategy which is used to effectively filter out noisy data, our model can be re-trained from higher-quality data adaptively and shows faster training speed and better performance. 
More detailed analysis of ECL strategy is conducted in Sec \ref{Ensemble Confident Learning}, which shows that the method of re-training from numerous noisy datasets to high-quality datasets can effectively improve the generalization ability and performance of the model.

\renewcommand{\arraystretch}{1.2}
\begin{table}[!htp]
  \centering
  \begin{tabular}{c||c|c|c|c}
    \hline
    Filtered rank & 0.7 & 0.8 & 0.9 & 0.99 \\
    \hline
    R@1 & 17.66 & 17.82 & 18.02 & 14.70
 \\
    \hline
  \end{tabular}
  \vspace{-2mm}
    \caption{\textbf{Impact of filtered rank.} As the filtered rank increases from 0.7 to 0.9, the R@1 improves. However, once the filtered rank is too high, the recall drops.}
  \label{filtered_values}
  \vspace{-6mm}
\end{table}

\subsubsection{Combined with MLM Task Training}\

In the pre-training process, we design two dataloaders for the image-text pairs and text data.
The batch size of image-text data and text data are 180 and 40 respectively. 
The text data is loaded similarly with BERT: the masking probability is set to 15\% for each token, and 
each masked token has a 20\% chance to be replaced by other tokens. 

EfficientCLIP is a multi-task pre-trained model, and its final performance is affected by the weights between contrastive learning task and MLM task. 
In our experiment, we set the weights of these two tasks to be proportional to their batch sizes.
If the batch size of text data is too large, it will impair our model's performance  on image-text retrieval tasks; 
if the batch size is too small, the introduction of  MLM task will hardly improve the model's performance. 
In order to verify the effect of the weight of MLM task on model's image-text retrieval ability, 
we calculate the f1\footnote{$f1= \frac{2 \cdot precision\cdot recall}{precision+   recall} $} 
score under different text batch sizes on the validation set. 
The results are shown in Table \ref{fig:batch-f1}.

\renewcommand{\arraystretch}{1.2}
\begin{table}[!htp]
  \centering
  \begin{tabular}{c||c|c|c|c}
    \hline
    Text batch size & 30 & 40 & 50 & 60 \\
    \hline
    Val f1 score & 0.452 & 0.460 & 0.444 & 0.442
 \\
    \hline
  \end{tabular}
  \vspace{-2mm}
    \caption{\textbf{Impact of the text batch size.} When the batch size is less than 40, the MLM task is helpful for improving the model's performance. But once the batch size is larger than 40, the greater the weight of MLM task, the model will reduce the learning ability on image-text retrieval tasks, and the performance of image-text retrieval will decrease.}
  \label{fig:batch-f1}
  \vspace{-2mm}
\end{table}

In order to make the model more focused on image-text retrieval tasks in the later stage, we choose to remove the MLM task after stopping data filtering.

\subsection{Adaptive Cleaning Data}
\label{Adaptive Cleaning Data}

\renewcommand{\arraystretch}{1.2}
\begin{table}[!th]
  \centering
  \begin{tabular}{c||c|c|c}
    \hline
     Dataset size & 300M & 200M & 100M \\
    \hline
    Bad case percent \% &28.0 &8.0  &1.0 \\
    \hline
    Good case percent \% &21.0 &47.0  &66.0
 \\
    \hline
  \end{tabular}
  \vspace{-2mm}
    \caption{\textbf{Performance of ECL to clean noise.} Dataset size, bad case percent and good case percent represent the size of the dataset we sample, the proportion of bad cases and good cases in the sampled data respectively.}
  \label{fig:size-bad}
  \vspace{-2mm}
\end{table}

\begin{figure*}[h!]
\centering
\includegraphics[width=1\textwidth]{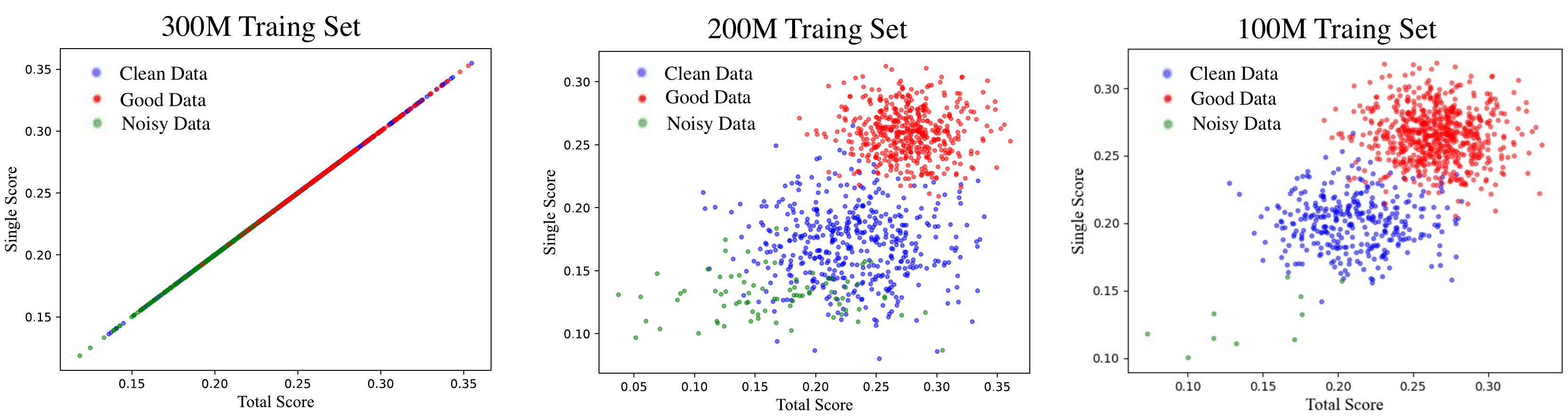}
\vspace{-7mm}
\caption{\textbf{Distribution of training set with different sizes.} The vertical axis and horizontal axis represent the score of the corresponding epoch and the total score, respectively. Good cases, clean cases, noisy cases represent the highly correlated pairs, correlated pairs, uncorrelated pairs (judge by human subjectivity), respectively. The first figure demonstrates the situation of first epoch where the total score equal to single epoch score.}
\label{fig:distribution}
\vspace{-6mm}
\end{figure*}

To verify the effect of our model for adaptive filtering, we randomly sample 1000 samples from the 300 million, 200 million, and 100 million datasets (200 million and 100 million datasets are filtered by ECL on the 300 million dataset) respectively, and label them manually. We count the proportion of bad cases and good cases in the samples and present it in Table \ref{fig:size-bad}. We also visualize the distribution of the manual data, as shown in Figure \ref{fig:distribution}. It can be seen that ECL has a strong ability to remove noisy data and separate the good cases from the other cases. 

\subsection{Comparison to State-of-the-art}
\label{comparison}

In this section, we evaluate the effectiveness (generalization and discriminative ability) of EfficientCLIP under different scenarios. To measure the capability of task-agnostic models, zero-shot evaluation have been widely adopted and proved as being much more representative of a model's ability~\cite{radford2021learning}. Thus, we conduct cross-modal retrieval in both Chinese and English datasets under zero-shot setting. To further illustrate the zero-shot transfer ability to downstream task, we also evaluate on single-modal tasks, such as text classification and text retrieval. The CLIP model used in this section is CLIP ViT-B/32 because our model is based on this model to show the text encoder's improvement by our approach.

\subsubsection{Cross-Modal Retrieval in Chinese}\

AIC-ICC~\cite{wu2017ai} is the only publicly-available Chinese multi-modal dataset. The training set contains 210,000 images and 1.05 million descriptions (5 captions for each image), and the validation set contains 30,000 images and 150,000 descriptions. We evaluate the zero-shot transfer ability on the test subset (10,000 data) following the same setting as WenLan~\cite{huo2021wenlan}. To compare with other benchmarks (mainly trained with English data), we translate Chinese text into English via Google translation API. We also distill CLIP into Chinese domain (details can be seen in Sec \ref{distillation}) to decrease the impact of translation, while UNITER (single tower) uses the translation directly.


Table \ref{table1} presents the cross-modal retrieval results. We can observe that
our EfficientCLIP significantly outperforms other benchmarks on both the text-to-image and image-to-text retrieval subtasks, showing the effectiveness of the proposed approach in this paper.

\renewcommand{\arraystretch}{1.2}
\begin{table}[H]
    \centering
    \scalebox{0.9}{
    

\begin{tabular}{l|ccc|ccc}
    \hline
    \multirow{2}*{Method} & \multicolumn{3}{c|}{Text2Image(\%)} & \multicolumn{3}{c}{ Image2Text(\%)} \\  
	\cline{2-7}  
    ~ & R1 & R5 & R10 & R1 & R5 & R10 \\
	\hline \hline
    CLIP$^{\dag}$ & 7.80 & 18.50 & 25.00 & 13.40 & 27.30 & 35.10 \\
    \hline
    UNITER$^{\dag}$ &9.80 & 23.30 & 31.40 & 14.80 & 29.80 & 37.90 \\
    \hline
    CLIP$^{\ddag}$ & 11.06 & 24.89  & 33.28 & 18.33 & 34.45 & 43.23 \\
    \hline
    WenLan &14.40 & 30.40 & 39.10 & 20.30 & 37.00 & 45.60 \\
    \hline
    E-CLIP &\textbf{18.02} & \textbf{36.33} & \textbf{45.91} & \textbf{26.33} & \textbf{44.73} & \textbf{52.95} \\
    \hline
    
\end{tabular}

    }
    \vspace{-2mm}
	\caption{\textbf{Evaluation results for the cross-modal retrieval tasks on the AIC-ICC  test subset.} $^{\dag}$ and $^{\ddag}$ means translation and distillation, respectively. E-CLIP represents the EfficientCLIP.}
	\vspace{-5mm}
	\label{table1}
\end{table}

\subsubsection{Cross-Modal Retrieval in English}\

Because of language differences which come from the different culture and style, transferring a model to another language will pose a huge challenge to the generalization of model. When CLIP transfers to Chinese domain, its performance is far lower than the current Chinese model~\cite{huo2021wenlan}.
So, to provide the more comprehensively empiricle evidence of EfficientCLIP's generalization ability, we directly compare with benchmark~\cite{radford2021learning} trained on English datasets. As the original EfficientCLIP is trained on Chinese datasets, we transfer the EfficientCLIP (text encoder) into English domain through cross-language knowledge distillation (details are the same as the distillation of Chinese-to-English but the student and teacher models are exchanged). A common image caption dataset in English is adopted for evaluation.

\textbf{COCO~\cite{veit2016coco}} is common adopted caption datasets. We use the COCO2014 test set (5000 images and 24716 captions) to evaluate our EfficientCLIP model. The results are shown in Table \ref{table2}.
 The distilled EfficientCLIP outperforms CLIP by nontrivial margin on text-to-image retrieval, while achieving competitive results on image-to-text retrieval tasks.

\renewcommand{\arraystretch}{1.2}
\begin{table}[H]
    \centering
    \scalebox{0.9}{
    

\begin{tabular}{l|ccc|ccc}
    \hline
    \multirow{2}*{Method} & \multicolumn{3}{c|}{Text2Image(\%)} & \multicolumn{3}{c}{ Image2Text(\%)} \\  
	\cline{2-7}  
    ~ & R1 & R5 & R10 & R1 & R5 & R10 \\
	\hline \hline 
    CLIP & 30.75 & 56.60 & 67.73 & \textbf{50.78} & 75.10 & 83.70 \\
    \hline
    E-CLIP$^{\ddag}$ &\textbf{35.12} & \textbf{61.71} & \textbf{72.77} & 50.70 & \textbf{76.56} & \textbf{84.98} \\
    \hline
    
\end{tabular}

    }
    \vspace{-2mm}
	\caption{\textbf{Evaluation results of the cross-modal retrieval tasks on the COCO2014 test set.}  $^{\ddag}$ mean distillation. E-CLIP represents the EfficientCLIP.}
	\vspace{-2mm}
	\label{table2}
\end{table}

\subsubsection{Single-Modal Evaluation}\
\label{nlp test set}

Previous cross-modal pre-trained models~\cite{radford2021learning,huo2021wenlan} usually cannot effectively adapt to single-modal (NLP) scenarios~\cite{li2020unimo}. We bridge this gap between cross-modal pre-training and its single-modal (NLP) counterpart by validating the generalization ability of EfficientCLIP on several NLP tasks. 


\textbf{(1) Text Classification.} To evaluate the NLU (Natural Language Understanding) capability on short texts, we adopt news classification task (TNEWS) in Chinese dataset CLUE~\cite{xu2020clue} as our benchmark. Each item in TNEWS has three attributes, NEWS ID, NEWS TYPE and NEWS TITLE. The training set contains 53,360 samples, and the test set and validation set both contain 10,000 samples. We compare with some popular benchmarks, such as XLNET~\cite{yang2019xlnet}, ALBERT-xxlarge~\cite{lan2019albert}, and ROBERTA-xxlarge on the validation set. The results are shown in Table \ref{table5}.

\renewcommand{\arraystretch}{1.2}
\begin{table}[H]
    \centering
    \scalebox{0.9}{

    

\begin{tabular}{l||c|c|c|c}

	\hline
    Model & XLNET & RoBERTa* & ALBERTA* & E-CLIP
    \cr\hline
    Acc(\%) &56.10 & 58.61 & 59.50 & \textbf{67.20}
    \cr\hline
    
\end{tabular}
    }
    \vspace{-2mm}
	\caption{\textbf{Results on TNEWS dataset.} * represents as  "-xxlarge". E-CLIP represents the EfficientCLIP.}
	\vspace{-4mm}
	\label{table5}
\end{table}

\textbf{(2) Text Retrieval.} To measure the discriminative ability of text embedding and the zero-shot transfer ability of facing unseen tasks, we evaluate on AIC-ICC~\cite{wu2017ai} test subset (same as Section 4.5.1, but only use the texts) where each image has 5 corresponding descriptions. We randomly select one of the 5 texts as the query, and the rest of 4 texts as the key. We reproduce SimCSE~\cite{gao2021simcse} and SIMBERT as our compared benchmarks (details can be found in supplementary materials Appendix \ref{reproduction of SimCSE}). For CLIP, we translate the Chinese text into English via Google translation API and use the distilled CLIP for comparison. The results can be found in Table \ref{table6}. We also evaluate on AFQMC and LCQMC~\cite{liu2018lcqmc} which are commonly used Chinese semantic matching datasets. The results can be found in Appendix \ref{other comparison}, Table \ref{table10}.

\renewcommand{\arraystretch}{1.2}
\begin{table}[t]
    \centering
    

\begin{tabular}{l|ccc}
    \hline
    \multirow{2}*{Method} & \multicolumn{3}{c}{Text Match(\%)} 
	\cr\cline{2-4}  
    ~ & R1 & R5 & R10
	\cr\hline\hline 
    CLIP$^{\dag}$ &22.94 & 36.25 & 42.65
    \cr\hline
    WenLan &31.44 & 45.32 & 52.98
    \cr\hline
    CLIP$^{\ddag}$ &37.34 & 53.26 & 60.64
    \cr\hline
    SimBERT &38.4 & 55.41 & 62.10
    \cr\hline
    SimCSE &39.64 & 56.49 & 63.24
    \cr\hline
    EfficientCLIP &\textbf{43.48} & \textbf{60.36} & \textbf{67.74}
    \cr\hline
    
\end{tabular}
    \vspace{-2mm}
	\caption{\textbf{Evaluation results for short text retrieval on AIC-ICC test subset.} $^{\dag}$ and $^{\ddag}$ means translation and distillation, respectively.}
	\vspace{-6mm}
	\label{table6}
\end{table}


We also provide the results of comparison on text retrieval task in English domain, which can be seen in Appendix \ref{text retrieval}.

As illustrated above, EfficientCLIP shows advantages over all other competitors on several common NLP tasks, which validates that extra single-modal pre-training enriches the discriminative power of single-modal (text) branch.

\subsection{Ablation Study}

In the section, we investigate the effects of single-modal pre-training and Ensemble Confident Learning (ECL) strategy on the overall performance on common benchmarks.

\subsubsection{Single-Modal Pre-Training}\
\label{nlp pre-train}


We evaluate the impact of the extra single-modal pre-training branch on cross-modal and single-modal tasks. The cross-modal and text retrieval tasks are conducted on AIC-ICC dataset, while the text classification is evaluated on TNEWS dataset.The results are shown in Table \ref{table8}.

As shown, the extra single-modal (text) pre-training branch not only enhances model's zero-shot transfer ability on text classification and text retrieval (single-modal) tasks, but also improves the performance on text-to-image retrieval (cross-modal) tasks. As expected, the performance gains from text retrieval (3.18\%$\uparrow$) and text classification (3.80\%$\uparrow$) are higher than cross-modal retrieval (0.75\%$\uparrow$).

\renewcommand{\arraystretch}{1.2}
\begin{table}[H]
    \centering


\begin{tabular}{l|c|c|c}

	\hline
    Models & T2I (R@1) & TC (Acc) & TR (R@1)
    \cr\hline\hline
    E-CLIP (wo) &17.27 & 63.40 & 40.30
    \cr\hline
    E-CLIP (w) &\textbf{18.02} & \textbf{67.20}& \textbf{43.48}
    \cr\hline
\end{tabular}
     \caption{\textbf{Effect of single-modal pre-training.} T2I, TC and TR denote text-to-image retrieval, text classification and text retrieval respectively. (w) and (wo) mean training with single-modal pre-training or not. E-CLIP is the EfficientCLIP.}
	\vspace{-4mm}
	\label{table8}
\end{table}

\subsubsection{Ensemble Confident Learning}\
\label{Ensemble Confident Learning}

The other key part in our approach is the Ensemble Confident Learning strategy, where we adaptively filter data for improving training efficiency and model performance. We train a series of four models under different settings and evaluate on the same tasks as Sec 4.6.1. The results are shown in Table \ref{table9}.

\renewcommand{\arraystretch}{1.2}
\begin{table}[H]
    \centering


\begin{tabular}{l|c|c|c}

	\hline
    Method & T2I & TC  & TR 
    \cr\hline\hline
    EfficientCLIP (300M) &10.77 & 65.20 & 32.70
    \cr\hline
    EfficientCLIP (200M*) &14.87 & 66.70 & 40.65
    \cr\hline
    EfficientCLIP (100M*) &16.72& 65.92 & 40.38
    \cr\hline
    EfficientCLIP+ECL (300M) &\textbf{18.02} & \textbf{67.20} & \textbf{43.48}
    \cr\hline
\end{tabular}
     \caption{\textbf{Effect of ECL strategy.} T2I, TC and TR denote text-to-image retrieval (R@1), text classification (Acc) and text retrieval (R@1) respectively. M stands for a million of uncleaned data. * means the data is filtered with ECL strategy.}
	\vspace{-4mm}
	\label{table9}
\end{table}

As the table \ref{table9} shown, as the quality of data increases, while the scale of data decreases, the retrieval performance still gains significant improvement, but NLP downstream  performance degrades. Second, instead of only using the highly correlated data (denoted with *), ECL strategy adopts an adaptive way to select training subset from huge noisy dataset, which is shown to be more generalized on both cross-modal and single-modal tasks.

\section{Conclusion}

In this work, we introduce an efficient cross-modal pre-training method called EfficientCLIP. We propose an Ensemble Confident Learning (ECL) strategy to adaptively filter the noisy dataset. We show the value of single-modal non-paired data for improving the generalization performance. We claim that EfficientCLIP achieves the SOTA performance on  Chinese cross-modal retrieval tasks, surpassing CLIP in English scene, and outperforms benchmarks on single-modal downstream tasks.

\appendix

\begin{figure*}[t]
\begin{center}
\includegraphics[width=1\textwidth]{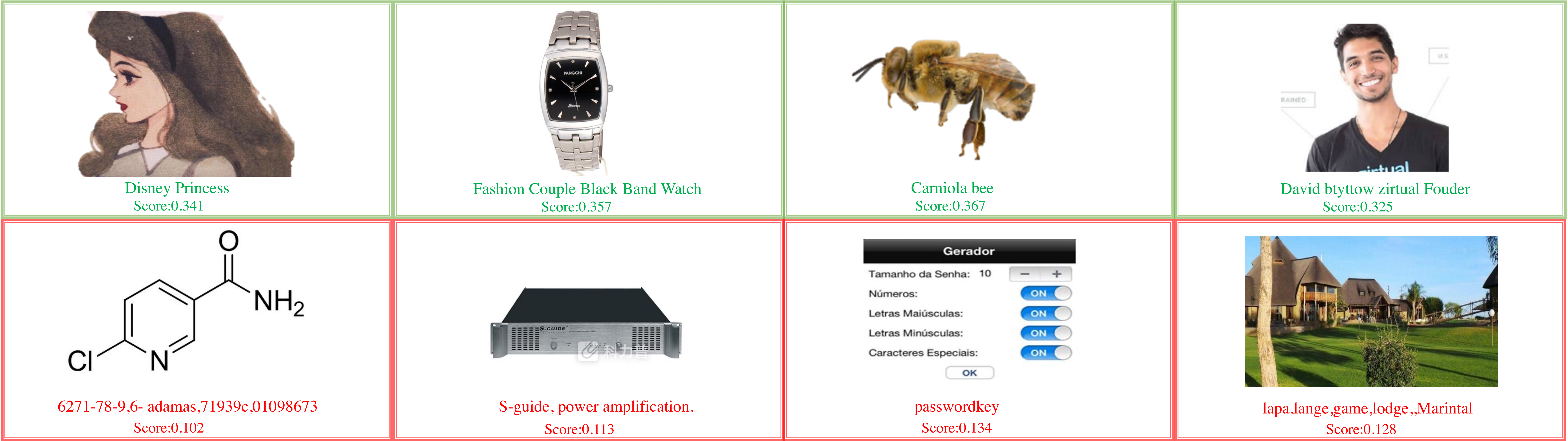}
\caption{\textbf{Example of raw text-image pairs on the internet} (The captions are translated from Chinese by Google Translation API). The high-quality pairs are indicated by \textcolor{green}{GREEN} (The first row of pictures), the noisy pairs are indicated by \textcolor{red}{RED} (The second row of pictures). The scores generated by our model shows semantic similarity for image-text pairs.}

\label{more case}
\end{center}
\vspace{-0.5cm}
\end{figure*}

\section{ Analyses for ECL}
\label{Explanation}

\subsection{Related Experients of ECL}\
\label{exp of ecl}
\vspace{-4mm}

 \begin{figure}[htp]
 \centering
 \includegraphics[width=0.5\textwidth]{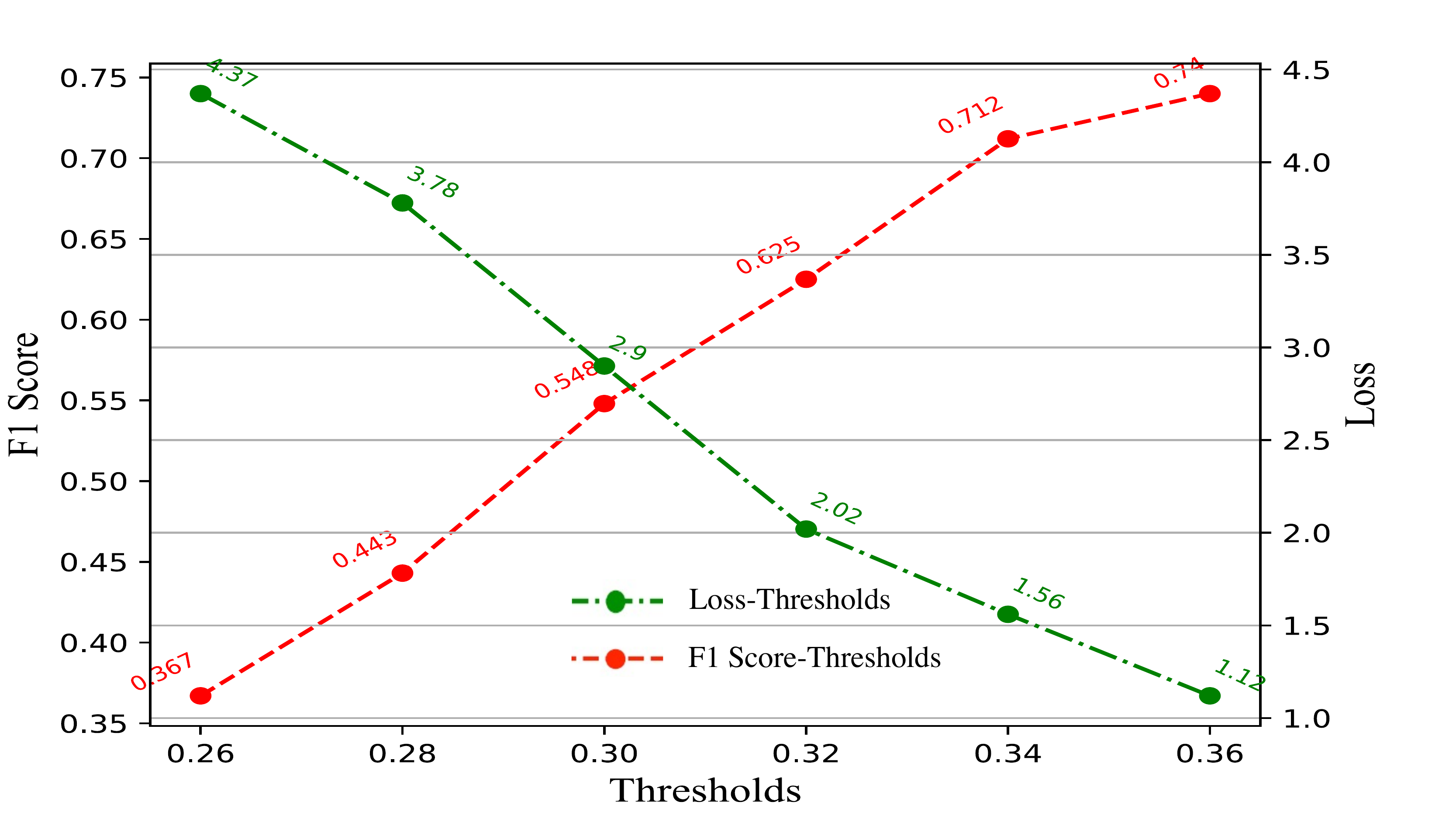}
 \vspace{-6mm}
 \caption{
 \textbf{Model performance on validation sets with different thresholds.} Higher f1 score or lower loss demonstrate the better performance of model in the validation sets.}
 \label{fig:image 2}
 \vspace{-6mm}
 \end{figure}

Highly correlated (high quality) image-text pairs contain rich and useful supervision which can be sufficiently learned by model. In contrast, the knowledge of noisy pairs are hard to be absorbed, which leads to the model's ability of discriminating the difference between high quality pairs and noisy pairs in the training. We conduct experiments to investigate this perspective. We train a cross-modal model through weakly supervised contrastive learning on noisy 300 million image-text pairs, and evaluate its performance on datasets of different qualities. We use the distillation model trained in Sec 3.1 to calculate the cosine similarity score of each training pair, and randomly select 20,000 pairs with cosine similarity score larger than the thresholds (we set to the table \ref{fig:threshold-recall}) as validation sets. We assume that the validation set with a higher threshold approximately represents higher quality. As illustrated in Figure \ref{fig:image 2}, the model performs better on the higher quality  (higher thresholds) validation sets. The result indicates that the model performs better on those high-correlated data even if it is trained on the noisy dataset, and shows the crucial difference between the model's performance on validation sets with different thresholds (qualities). Thus, the above result provides a powerful evidence of the training model's ability to separate the high quality pairs from the noisy pairs.  

 \renewcommand{\arraystretch}{1.2}
 \begin{table}[!htp]
   \centering
   \begin{tabular}{c|c|c|c|c}
      \hline
       Threshold & 0.28 & 0.30 & 0.32 & 0.34 \\
      \hline\hline
      R@1 & 10.87 & 11.45 & 12.34 & 12.66
   \\
      \hline
   \end{tabular}
   \vspace{-2mm}
      \caption{\textbf{Impact of the setting of thresholds.} The R@1 represents Recall@1 (on the AIC-ICC text-to-image task) of models finetuned with corresponding datasets.}
   \label{fig:threshold-recall}
   \vspace{-2mm}
  \end{table}

 In addition, to confirm that the similarity scores calculated by the distillation model are positively correlated with the ground truth similarity, we finetune the distillation model on those datasets of different thresholds (each dataset is composed of 100,000 randomly selected pairs). We evaluate the finetuned model on text-to-image task of the AIC-ICC~\cite{wu2019large} test set. The results are shown in Table \ref{fig:threshold-recall} (because the number of data of the datasets with higher thresholds are lower than 100,000, so the highest threshold used is 0.34). We find that the higher threshold is set to filter data, the higher R@1 score on the test set. It can be explained that because the model is finetuned on a higher quality dataset with higher threshold, the finetuned model will perform better on the test set. So, as illustrated in above analysis, what the similarity scores produced by the distillation model are positively correlated with the correlation of image-text pairs can be verified.

\subsection{Effectiveness of ECL}\
\label{eff of ecl}
\vspace{-4mm}

Regarding the three-step process of Sec 3.2, we have two considerations: 
(1) A motivation to use the distillation model as the initial model is not only to save training cost, but also to endow the model with initial ability to judge data distribution. Thus, our model can give a score with high confidence even at the beginning. 
(2) If we only take the score of the distillation model as the total score, it is easy to filter out the strongly related image-text pairs with low scores. In this way, the final model is easy to converge to the local optimum relying on the distillation model. Therefore, we propose to ensemble scores of multiple models through exponential smoothing. The details of the ablation experiment can be found in Appendix \ref{Ensemble Scoring Shadow Models}.

Before explaining the ECL, we first give a priori $Q$ (scoring shadow model), and set the filtered rank as $\lambda$ ($0<\lambda<1$). Based on the prior $Q$, we can estimate the distribution of the data, and after re-ranking, filter out the data ranking after $\lambda$. We set the number of  clean data as $n$, the noisy data as $m$. After filtering, the number of  clean data and noisy data change to n* and m* respectively. Then, this process can always satisfy $\frac{n^*}{n} >\frac{m^*}{m} $. Related experiments can be seen in Sec 4.4.

Let us give an example: we assume that there are $(n+m)$ dog images $I_d$ in the dataset, while only $n$ images  correspond to the dog descriptions $D_d$ (the dataset represents as $\{D_d \cap I_d\}$), and the $m$ images correspond to the noisy descriptions $D_n$ (the dataset represents as $\{D_n \cap I_d\}$). Our ideal optimization function is $W^*=\underset{W}{\arg\max} P(D_d|I_d, W)$. However, based on the Monte Carlo method, the radio of sampling data of $\{D_d \cap I_d\}$ to $\{D_n \cap I_d\}$ in the training is $n:m$. So the real optimization function can be approximately represented as  $W^*=\underset{W}{\arg\max} (P(D_d|I_d,W)+\frac{m}{n}P(D_n|I_d,W))$ because of the training with noisy data.

We assume that after the first round of filtering, the number of dataset $\{D_d \cap I_d\}$ is $n^*=v_1\times n(0<v_1<1)$, and dataset $\{D_n \cap I_d\}$ is $m^*=u_1\times m(0<u_1<1)$, and according to the prior, there always satisfies $\frac{n^*}{n} >\frac{m^*}{m}$ (which also satisfies $v_1>u_1$). Then, in the next epoch, the optimization function can be expressed as $W^*=\underset{W}{\arg\max} (P(D_d|I_d,W)+\frac{m}{n}\frac{u_1}{v_1 } P(D_n|I_d,W))$. In this iteration, after $K$ epochs, the optimization function is $W^*=\underset{W}{\arg\max} (P(D_d|I_d,W)+\frac{m\prod_{1}^{K} u_i}{n\prod_{1}^{K}v_i} P(D_n|I_d,W))$. We denote $\frac{\prod_{1}^{K}u_i}{\prod_{1}^{k}v_i}$ ($0<\frac{\prod_{1}^{K}u_i}{\prod_{1}^{k}v_i}<1$) as a regular term, then we can regard ECL as a regularization process. As the ECL is an iterative method, we can control the regular term to the most suitable extent for the best model's performance. ECL greatly reduces the impact of the noisy sample during training, making the model more accurate to predict for input from the real world. Related experiments can be seen in Sec 4.6.2 and Sec 4.3.

\section{Memory Queue}
\label{memory}

Contrastive learning methods heavily depend on the number and quality of negative samples for generating high-quality representations that are robust to noise interference and beneficial to performance improvement. 
In order to free negative samples from the constraints of batch size, we introduce a memory queue to expand the number of negative samples. 

In MOCO~\cite{he2020momentum}, a momentum-based updating rule is designed for the key encoder to ensure the consistency of the dictionary of a queue. 
However, when the size of queue is too large, it is hard to maintain consistency due to the parameters update of key encoder which is not sync with the query encoder. 
The discrepancy of negative samples' features might result in training instability and performance degradation~\cite{he2020momentum}.
To alleviate the problem, we utilize the frozen image encoder as the key encoder and only train the text encoder. 
Since the keys in memory queue all come from the same image encoder, there is no inconsistency no matter how large the queue is. 
The contrastive loss in our method is formulated as following:

\begin{equation}
\underset{x, x^{+}, x^{-}}{\mathbb{E}}\left[-\log \left(\frac{e^{f(x)^{T} f\left(x^{+}\right)}}{e^{f(x)^{T} f\left(x^{+}\right)}+e^{f(x)^{T} f\left(x^{-}\right)}}\right)\right].
\label{fig:loss}
\end{equation}  
where $(x, x^+)$ and $(x, x^-)$ are positive and negative pairs, respectively. $f(x)$ represents the encoder forward function.

\section{Creating Datasets}
\label{Datasets}

\subsection{Public Datasets}\

\textit{Chinese-English Text Pairs}. To conduct cross-language knowledge distillation, we collect Chinese-English paired data from AI Challenge Machine Translation\footnote{https://challenger.ai}(denoted as AIC), WMT20\footnote{http://www.statmt.org/wmt20/}, and other public Chinese-English translation datasets, totaling 80 million translation pairs. 

\textit{Chinese Text Data}. For the extra single-modal branch, we adopt the single-modal text dataset from CLUE~\cite{xu2020clue}, which is the largest Chinese language understanding evaluation benchmark. We clean the dataset by removing data with low Chinese character ratio (50\% in our case) and meaningless symbols. We finally get a Chinese text dataset of 56G size, totaling 20,329,263 documents.

\subsection{ Web-crawled Datasets}\
 
\textit{Image-Text Pairs for training}.
To construct a large scale Image-Text dataset for contrastive learning, we establish a Chinese word dictionary including 4 million Chinese vocabulary. Each word in the dictionary is used as query to crawl image-text pairs from Chinese Search Engine (Baidu Pictures and Baidu Baike). We simply clean the raw crawled pairs as we did for text data and get a Chinese image-text paired training dataset of 300 million pairs. Details of preprocessing of the training set can be found in Appendix \ref{data processing}.

\textit{Image-Text Pairs for validation}. 
To verify the effectiveness of our method in time, we extra collect image-text pairs from various scenarios on the internet. We collect image-text pairs from Baidu Pictures, Baidu Baike, Toutiao, hashtag, and other sources. Specifically, we first use the distillation model to calculate the cosine similarity score for each image-text pair. Based on these scores, we sort the dataset and take out the first 1 million of data for next cleaning stage.Then, in order to extract more general and common data, we extract the data containing common words defined in a 40 thousand entity vocabulary which is collected from GitHub. 
Finally, we clean out the top-10,000 (sorted by the cosine similarity score of pairs) image-text pairs as our validation set, which is not covered by the training set.The details of sizes of these datasets can be found in Table \ref{datasets}.

\renewcommand{\arraystretch}{1.2}
\begin{table}[t]
  \centering
  \begin{tabular}{c||c|c|c|c}
    \hline
    Dataset Name & CET & TD & IMT & IMV \\
    \hline
    number of data & 80M & 20.3M & 300M & 10K \\
    \hline
    sizes & 14.2G & 56G & / & / \\
    \hline
  \end{tabular}
    \caption{\textbf{Details of used datasets.} CET, TD, IMT, IMV represent Chinese-English Translation, Text Data, Image-Text Training Set and Validation Set respectively. M, K represent a million and a thousand respectively.}
    \vspace{-5mm}
  \label{datasets}
\end{table}

\section{Data Processing}
\label{data processing}

In terms of the preprocessing of training set, we try a variety of methods, and prove that cleaning text branch of image-text pairs has a considerably significant improvement on the performance of model. We first train on the raw 300 million data, while the R@1 of trained model on text-to-image task of AIC-ICC test set is just 6.58 which is far lower than the current SOTA~\cite{huo2021wenlan}. Therefore, we try to perform preprocessing for text and compare the effects of various cleaning methods on model's performance.

First, we remove the redundant HTML, space symbols (such as “…”  “---”) and all emojis, which are meaningless in text. We notice that the crawled text data exists enormous interval symbols such as  “\&”, “-”, etc. We replace these interval symbols with “,” to make sentences more cleaned and fit with general data. Although all we crawl are Chinese websites, some English sentences and extremely short sentences will inevitably appear in text data. Therefore, we perform rules with a Chinese character ratio of less than 0.5 to remove English sentences, and  a length which is less than 4 to remove short sentences. We conduct related experiments to explore the effects of removing these two kinds of sentences on the performance of model. We train models with same hyperparameters and regular contrastive learning method on the raw 300 million data, the removal of English sentence, short sentence data, and the data removing two kinds of sentences, respectively. And we use the image-text retrieval tasks on AIC-ICC for evaluation. The details can be found in Table \ref{table12}.

\renewcommand{\arraystretch}{1.2}
\begin{table}[H]
    \centering
    

\begin{tabular}{l|ccc|ccc}
    \hline
    \multirow{2}*{Method} & \multicolumn{3}{c|}{Text2Image(\%)} & \multicolumn{3}{c}{ Image2Text(\%)} \\  
	\cline{2-7}  
    ~ & R1 & R5 & R10 & R1 & R5 & R10 \\
	\hline \hline
    No Clean & 6.58 & 15.19 & 20.89 & 12.14 & 25.23 & 33.20 \\
    \hline
    C\&S &10.29 & 22.98 & 30.74 & 20.73 & 37.27 & 46.00 \\
    \hline
   C\&S\&E & 10.56 & 23.39  & 31.20 & 21.15 & 37.49 & 46.00 \\
    \hline
    C\&E &\textbf{10.77} & \textbf{23.99} & \textbf{31.86} & \textbf{21.50} & \textbf{38.30} & \textbf{46.83} \\
    \hline
\end{tabular}

     \caption{\textbf{Impact of data cleaning method.} C\&S, C\&S\&E, C\&E represent cleaning short text, cleaning short text and English text, cleaning English text respectively.}
	\vspace{-2mm}
	\label{table12}
\end{table}

The above results suggest that when all English sentences are removed, the performance of model on the test set is optimal. So we remove all English sentences in the dataset, and retain extremely short sentences.

\begin{figure}[H]
\begin{center}
\centerline{\includegraphics[width=1.0\columnwidth]{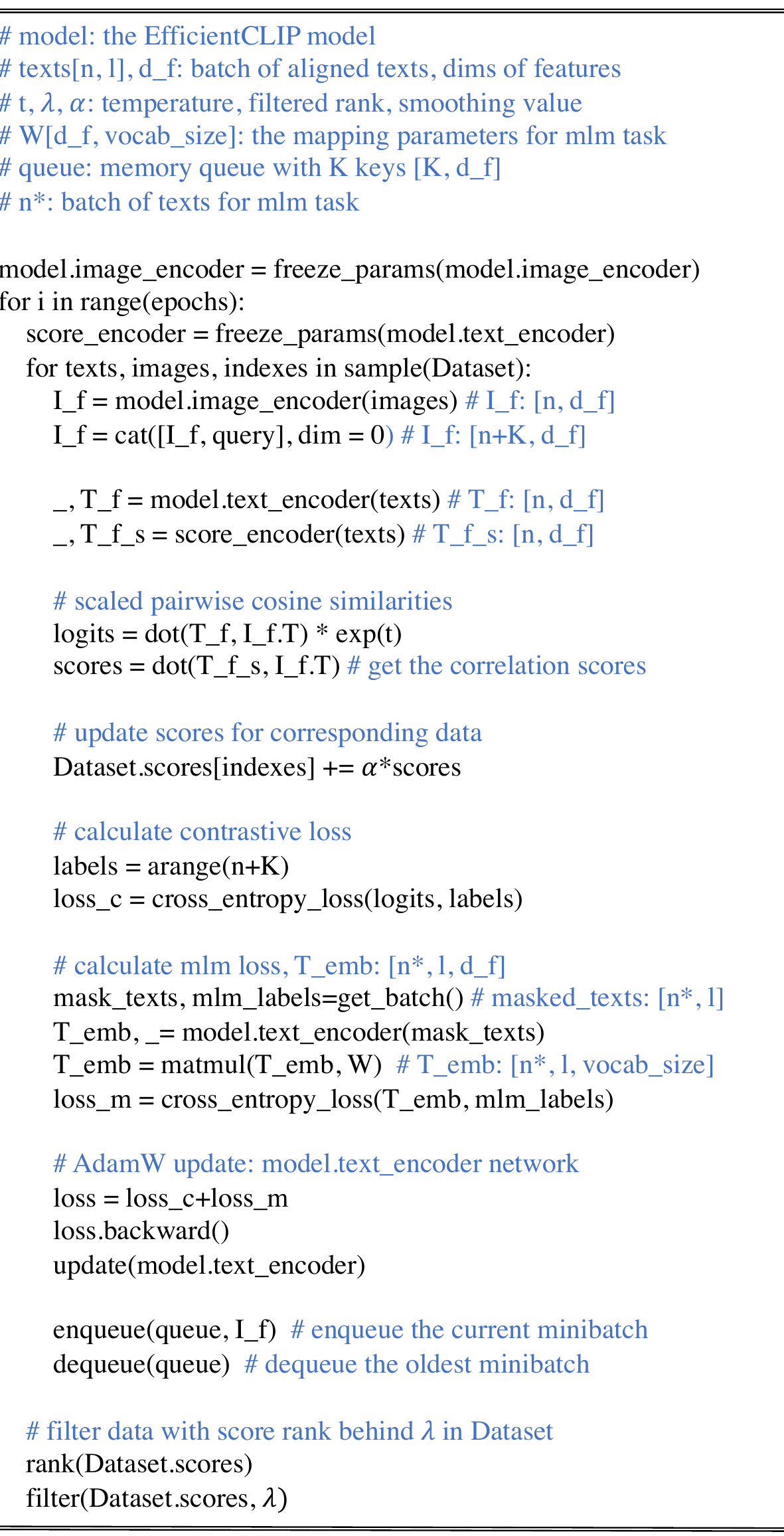}}
\caption{\textbf{Pseudocode for an implementation of EffientCLIP.} The text encoder have two outputs, including the embeddings of tokens (the first output) and the features of sentences (the second output).}
\label{pseudocode}
\end{center}
\vspace{-1em}
\end{figure}

\section{ Reproduction of SimCSE}
\label{reproduction of SimCSE}

For reproducing SimCSE to a great extent, we refer to the hyperparameters in SimCSE paper~\cite{gao2021simcse} and utilize 10 million sentences collected from Chinese Wikipedia (for unsupervised) and the BQ Corpus~\cite{chen2018bq}  dataset (for supervised) to reproduce. In terms of data preprocessing, we choose to divide the text into multiple shorter texts with a length of less than 77 as input data (the same as EfficientCLIP), and use text data processing method the same as Appendix \ref{data processing} to perform simple data cleaning. For better performance, we choose Roberta case (the best model mentioned in SimCSE paper) as the backbone of our SimCSE model. We add [CLS] token to each text, and use the representation at the position of [CLS] token as its sentence embedding. In the experiment, we set the learning rate to 1e-5 and the weight decay to 0.0001. For selection of dropout, we refer to the setting of SimCSE paper and choose the default 0.1. We use dropout noise as data augmentation to create positive samples of text, and use other randomly selected texts as negative samples for contrastive learning. We refer to the recommendation of SimCSE, where we add Masked Language Model task as auxiliary loss, which helps model avoid catastrophic forgetting of token-level knowledge.
For the purpose of speeding up the training of model, we also leverage MMAP method to load data and use half-precision optimization based on Deepspeed~\cite{rasley2020deepspeed} framework to accelerate distributed training. Finally, we cost 24 GPU-days on V100 to get a Chinese SimCSE model. After unsupervised training, we finetune the model on the BQ Corpus for supervised learning to obtain the best performance.

We also explore more data augmentation methods to enhance the training effect of model. Because the efficiency of calling Google Translate API is not high as our expectation, we choose to use the Fairseq~\cite{ott2019fairseq} framework to train a Chinese-to-English model and an English-to-Chinese model for offline back translation. We perform simple data augmentation through back translation to create a positive sample of text, and use the same training method as SimCSE to reproduce SimBERT.

\section{Speeding up Training}
\label{speed}

To reduce the cost of pre-training, we use the distillation model as a coarse initialization. 
Further, in order to load data with less time and memory cost, we utilize the image encoder of CLIP ViT-B/32 to extract image features before training and store them in a binary file. 
While using image features, we utilize the corresponding pointer to quickly extract the features by MMAP (A method for disk mapping memory).

\renewcommand{\arraystretch}{1.2}
\begin{table}[!htp]
  \centering
  \begin{tabular}{c|c|c}
    \hline
    Hyperparameters & E-CLIP-small & E-CLIP \\
    \hline\hline
    Number of Layers & 12 & 32
 \\
    \hline
    Hidden Size & 512 & 512
 \\
    \hline
    Dropout & 0.1 & 0.1
 \\
    \hline
    Attention Dropout & 0.1 & 0.1
 \\
    \hline
    Weight Decay & 0.0001 & 0.0001
 \\
    \hline
    number of parameters & 151M & 225M
 \\
    \hline
    R@1 & 16.82 & 18.02
 \\
 
    \hline
  \end{tabular}
    \caption{\textbf{Hyperparameters of two EfficientCLIP models.} R@1 represents the R@1 score obtained by the model on the text-to-image task of AIC-ICC, and E-CLIP represents the EfficientCLIP.}
  \label{hyperparameters}
\end{table}

\section{Other Comparison of EfficientCLIP}
\label{other comparison}




\subsection{ Text Retrieval in English}\
\label{text retrieval}

To further validate the discriminative ability of EfficientCLIP on text retrieval tasks, we also evaluate on COCO2014 test set (only use the captions). The results can be found in Table \ref{table7}.

\renewcommand{\arraystretch}{1.2}
\begin{table}[H]
    \centering
    

\begin{tabular}{l|c|c|c}
    \hline
    \multirow{2}*{Method} & \multicolumn{3}{c}{Text Match(\%)} 
	\cr\cline{2-4}  
    ~ & R1 & R5 & R10
	\cr\hline\hline 
    CLIP &38.34 & 59.44 & 68.52
    \cr\hline
    EfficientCLIP &\textbf{40.40} & \textbf{62.20} & \textbf{71.80}
    \cr\hline
    

\end{tabular}
	\caption{\textbf{Evaluation results for short text retrieval on COCO2014 test set.}}
	\vspace{-2mm}
	\label{table7}
\end{table}

\renewcommand{\arraystretch}{1.2}
\begin{table}[H]
    \centering
    \scalebox{0.9}{
    

\begin{tabular}{l|ccc|ccc}
    \hline
    \multirow{2}*{Method} & \multicolumn{3}{c|}{AFQMC(\%)} & \multicolumn{3}{c}{LCQMC(\%)}  
	\cr\cline{2-7}  
    ~ & R1 & R5 & R10 & R1 & R5 & R10
	\cr\hline\hline 
    CLIP$^{\dag}$ &6.43 & 14.80 & 19.96 & 57.17 & 78.22 & 82.74
    \cr\hline
    WenLan &9.72 & 18.68 & 23.92 & 66.27 & 87.01 & 90.82
    \cr\hline
    CLIP$^{\ddag}$ &12.03 & 24.59 & 31.54 & 74.53 & 93.50 & 96.18
    \cr\hline
    SimBERT &11.23 & 25.10 & 30.98 & 76.10 & 93.23 & 96.44
    \cr\hline
    SimCSE &13.34 & 27.22 & 33.78 & 78.56 & 95.52 & 97.32
    \cr\hline
    E-CLIP &\textbf{15.77} & \textbf{30.72} & \textbf{36.54} & \textbf{81.58} & \textbf{98.10} & \textbf{98.88}
    \cr\hline
    
\end{tabular}
    }
	\caption{\textbf{Evaluation results for short text retrieval on AFQMC and LCQMC.} $^{\dag}$ and $^{\ddag}$ means translation and distillation, respectively. E-CLIP represents the EfficientCLIP.}
	\vspace{-2mm}
	\label{table10}
\end{table}

\section{Ensemble Scoring Shadow Models}
\label{Ensemble Scoring Shadow Models}

To provide a powerful evidence for the effectiveness of integrating multiple models at different epochs during filtering, we train a model by only using the distillation model as the scoring shadow model. The results are shown in Table \ref{table11}.

\renewcommand{\arraystretch}{1.2}
\begin{table}[H]
    \centering


\begin{tabular}{l|c|c|c}

	\hline
    Models & T2I (R@1) & TC (Acc) & TR (R@1)
    \cr\hline\hline
    E-CLIP (wo) &17.03 & 67.20 & 42.04
    \cr\hline
    E-CLIP (w) &\textbf{18.02} & \textbf{67.20}& \textbf{43.48}
    \cr\hline
\end{tabular}
     \caption{\textbf{Effect of ensemble scoring shadow models.} T2I, TC and TR denote text-to-image retrieval, text classification and text retrieval respectively. E-CLIP represents the EfficientCLIP.}
	\vspace{-2mm}
	\label{table11}
\end{table}

\newpage
\bibliography{aaai22}

\end{document}